\documentclass[pmlr,onecolumn,breaklinks,10pt]{jmlr}

\usepackage{longtable}

\usepackage{booktabs}

\usepackage{siunitx} 
\usepackage{cuted}
\usepackage{soul}
\usepackage{cleveref}

\usepackage{siunitx}
\usepackage{float}

\theorembodyfont{\upshape}
\theoremheaderfont{\scshape}
\theorempostheader{:}
\theoremsep{\newline}

\usepackage[nolist]{acronym}

\makeatletter
\newcommand*{\org@overidelabel}{}
\let\org@overridelabel\@verridelabel
\@ifpackagelater{acronym}{2015/03/21}{
  \renewcommand*{\@verridelabel}[1]{%
    \@bsphack
    \protected@write\@auxout{}{\string\AC@undonewlabel{#1@cref}}%
    \org@overridelabel{#1}%
    \@esphack
  }%
}{
  \renewcommand*{\@verridelabel}[1]{%
    \@bsphack
    \protected@write\@auxout{}{\string\undonewlabel{#1@cref}}%
    \org@overridelabel{#1}%
    \@esphack
  }%
}
\makeatother

\usepackage{url}

\usepackage{hyperref}

\usepackage[T1]{fontenc}

\title[Recent Advances, Applications and Open Challenges from ML4H Roundtables]{Recent Advances, Applications and Open Challenges in Machine Learning for Health: Reflections from  Research Roundtables at ML4H 2024 Symposium}

\author{
    \Name{Amin Adibi\nametag{\textsuperscript{1,2}}}\Email{amin.adibi@ubc.ca}\\
    \Name{Xu Cao\nametag{\textsuperscript{1}}} \Email{xucao2@illinois.edu}\\
    \Name{Zongliang Ji\nametag{\textsuperscript{1}}} \Email{jerryji@cs.toronto.edu}\\
    \Name{Jivat Neet Kaur\nametag{\textsuperscript{1}}} \Email{jnkaur@berkeley.edu}\\
    \Name{Winston Chen\nametag{\textsuperscript{1,}\textsuperscript{2}}} 
    \Email{chenwt@umich.edu}\\
    \Name{Elizabeth Healey\nametag{\textsuperscript{1,2}}} 
    \Email{ehealey@mit.edu}\\
    \Name{Brighton Nuwagira\nametag{\textsuperscript{2}}} 
    \Email{brighton.nuwagira@utdallas.edu}\\
    \Name{Wenqian Ye\nametag{\textsuperscript{2}}} 
    \Email{wenqian@virginia.edu}\\
    \Name{Geoffrey Woollard\nametag{\textsuperscript{2}}} \Email{geoffwoollard@gmail.com}\\
    \Name{Maxwell A Xu\nametag{\textsuperscript{2}}} 
    \Email{maxu@illinois.edu}\\
    \Name{Hejie Cui\nametag{\textsuperscript{2}}} 
    \Email{hejie.cui@stanford.edu}\\
    \Name{Johnny Xi\nametag{\textsuperscript{2}}} 
    \Email{johnny.xi@stat.ubc.ca}\\
    \Name{Trenton Chang\nametag{\textsuperscript{2}}} 
    \Email{ctrenton@umich.edu}\\
    \Name{Vasiliki Bikia\nametag{\textsuperscript{2}}} 
    \Email{bikia@stanford.edu}\\
    \Name{Nicole Zhang\nametag{\textsuperscript{2}}} 
    \Email{xi.zhang@mila.quebec}\\
    \Name{Ayush Noori\nametag{\textsuperscript{2}}} 
    \Email{anoori@college.harvard.edu}\\
    \Name{Yuan Xia\nametag{\textsuperscript{2}}} 
    \Email{lily.yuanxia@stat.ubc.ca}\\
    \Name{Md. Belal Hossain\nametag{\textsuperscript{2}}} 
    \Email{belal.hossain@ubc.ca}\\
    \Name{Hanna A. Frank\nametag{\textsuperscript{2}}} 
    \Email{hanna.frank@ubc.ca}\\
    \Name{Alina Peluso\nametag{\textsuperscript{2}}} 
    \Email{pelusoa@ornl.gov}\\
    \Name{Yuan Pu\nametag{\textsuperscript{2}}} 
    \Email{yuan.pu@yale.edu}\\
    \Name{Shannon Zejiang Shen\nametag{\textsuperscript{2}}} 
    \Email{zjshen@mit.edu}\\
    \Name{John Wu\nametag{\textsuperscript{2}}} 
    \Email{johnwu3@illinois.edu}\\
    \Name{Adibvafa Fallahpour\nametag{\textsuperscript{2}}} 
    \Email{adibvafa.fallahpour@mail.utoronto.ca}\\
    \Name{Sazan Mahbub\nametag{\textsuperscript{2}}} 
    \Email{smahbub@cs.cmu.edu}\\
    \Name{Ross Duncan\nametag{\textsuperscript{2}}} 
    \Email{rduncan@healthresearchbc.ca}\\
    \Name{Yuwei Zhang\nametag{\textsuperscript{2}}} 
    \Email{yz798@cam.ac.uk}\\
    \Name{Yurui Cao\nametag{\textsuperscript{2}}} 
    \Email{yuruic2@illinois.edu}\\
    \Name{Zuheng Xu\nametag{\textsuperscript{2}}} 
    \Email{zuheng.xu@stat.ubc.ca}\\
    \Name{Michael Craig\nametag{\textsuperscript{3}}} \Email{michael@valencelabs.com}\\
    \Name{Rahul G. Krishnan \nametag{\textsuperscript{3}}} \Email{rahulgk@toronto.edu}\\
    \Name{Rahmatollah Beheshti\nametag{\textsuperscript{3}}} 
    \Email{rbi@udel.edu}\\
    \Name{James M. Rehg\nametag{\textsuperscript{3}}} 
    \Email{jrehg@illinois.edu}\\
    \Name{Mohammad Ehsanul Karim\nametag{\textsuperscript{3}}} 
    \Email{ehsan.karim@ubc.ca}\\
    \Name{Megan Coffee\nametag{\textsuperscript{3}}} 
    \Email{megan.coffee@gmail.com}\\
    \Name{Leo Anthony Celi\nametag{\textsuperscript{3}}} 
    \Email{lceli@mit.edu}\\
    \Name{Jason Alan Fries\nametag{\textsuperscript{3}}} 
    \Email{jfries@stanford.edu}\\
    \Name{Mohsen Sadatsafavi\nametag{\textsuperscript{3}}}
    \Email{mohsen.sadatsafavi@ubc.ca}\\
    \Name{Dennis Shung\nametag{\textsuperscript{3}}} \Email{dennis.shung@yale.edu}\\
    \Name{Shannon McWeeney\nametag{\textsuperscript{3}}} \Email{mcweeney@ohsu.edu}\\
    \Name{Jessica Dafflon\nametag{\textsuperscript{1}}} \Email{jessica@valencelabs.com}\\
    \Name{Sarah Jabbour\nametag{\textsuperscript{1}}} \Email{sjabbour@umich.edu}\\
    \addr \textsuperscript{1} Organizing committee for ML4H 2024 Research Roundtables, \textsuperscript{2}Junior Chairs, \textsuperscript{3}Senior Chairs
}

\begin{document}

\maketitle

\section{Introduction}
\label{sec:intro}

The 4th Machine Learning for Health (ML4H) symposium was held in person on December 15-16, 2024, in the traditional, ancestral, and unceded territories of the Musqueam, Squamish, and Tsleil-Waututh Nations in Vancouver, British Columbia, Canada. The symposium included research roundtable sessions to foster discussions between participants and senior researchers on timely and relevant topics for the ML4H community~\citep{jeong2024recent,hegselmann2023recent}.

The organization of the research roundtables at the conference involved 13 senior and 27 junior chairs across 13 tables. Each roundtable session included an invited senior chair (with substantial experience in the field), junior chairs (responsible for facilitating the discussion), and attendees from diverse backgrounds with interest in the session’s topic.

\section{Organization Process}
\label{sec:organizationprocess}

We began by identifying an initial set of topics from articles published in the ML for health literature over the past 3-5 years, complemented by suggestions from ML4H roundtable subchairs. After eliminating duplicates, we curated a list of 24 candidate topics. Then, the roundtable subchairs vote for each topic and finalize the 13 topics. For each topic, we invited senior chairs with domain expertise, targeting one chair per roundtable. We then selected junior chairs, prioritizing two individuals with prior experience relevant to the topic. Researchers in the program committee are also welcomed to serve as junior chairs in the roundtable.

Before the event, junior and senior chairs collaborated to draft an introductory paragraph, which was shared on the ML4H website. They also proposed up to four discussion questions to guide the sessions. Given the popularity of roundtables from previous years, where two 25 min sessions were held, we instead had two 50-minute discussions, which allowed the participants to discuss topics in more depth. Following the event, the chairs submitted written summaries that highlight the key insights and takeaways from their discussions.

\section{Research Roundtables}

\label{sec:roundtables}
\subsection{Foundation Models and Multimodal AI}
The convergence of multimodal foundation models and healthcare presents a frontier where diverse data streams—from medical imaging and clinical text to biosignals and genomic data—intersect to create comprehensive patient representations. During this roundtable, participants explored critical questions shaping the development of multimodal AI in healthcare:
How does multimodality affect model robustness and feature learning?
What are the privacy and security implications of multimodal healthcare models?
How should demographic and clinical features be incorporated into foundation models?
How do we handle incomplete and misaligned data across modalities?
How do we address causality and confounding in multimodal healthcare models?
What are the trade-offs between general-purpose and healthcare-specific foundation models?
How do we effectively handle temporal data and evaluation in clinical settings?

\paragraph{Chairs:}
\textit{Jason Fries, Max Xu, and Hejie Cui} 

\noindent Special acknowledgment goes to Junyi Gao for helping with taking the notes. 

\paragraph{Background:}
The emergence of multimodal foundation models marks a pivotal advancement in artificial intelligence, particularly for healthcare applications where patient data naturally spans multiple modalities. These models transcend traditional single-modality approaches by simultaneously processing and integrating diverse data types—medical imaging, clinical narratives, structured EHR data, genomic information, and temporal biosignals. This multimodal capability offers unprecedented opportunities to capture the intricate interplay between different aspects of patient health, potentially leading to more comprehensive and nuanced clinical understanding.

Healthcare presents a unique proving ground for multimodal AI, where the synthesis of diverse data types is not just beneficial but often essential for accurate diagnosis and treatment planning. A clinician's assessment typically involves integrating visual information from medical imaging, structured data from lab tests, temporal patterns from vital signs, and contextual information from clinical notes. Multimodal foundation models aim to mirror and augment this holistic approach to patient care.

However, the healthcare domain poses distinct challenges for multimodal AI development. Beyond common hurdles of data privacy and model interpretability, healthcare applications must contend with temporal misalignment between different data streams, varying data quality across modalities, missing modalities in real-world settings, and the need to maintain clinical validity across all data types. This roundtable discussion aimed to explore these intricate dynamics, bringing together experts to examine how multimodal foundation models can be effectively developed and deployed in healthcare while addressing these unique challenges.

\paragraph{Discussion:}
The roundtable discussion explored key challenges and opportunities in developing multimodal foundation models for healthcare applications. Through our discussion, five major themes emerged, spanning technical, clinical, and practical considerations. These themes highlight the complexities of integrating multiple modalities in healthcare AI while maintaining clinical utility and practical feasibility. 

\subsubsection{Data Integration and Robustness}
The first topic highlighted how the integration of multiple data modalities presents both opportunities and challenges for healthcare AI. Participants noted that combining different data types (imaging, text, biosignals) could inherently increase model robustness by reducing reliance on spurious correlations and encouraging more grounded learning. Vision-language models were frequently cited as examples where multimodal integration leads to better generalization. However, this benefit comes with increased complexity in handling data alignment, as healthcare data is often incomplete across modalities. The group emphasized that while more modalities can enrich understanding, they can also limit the available sample size of fully aligned data sets.

\subsubsection{Clinical Context and Data Quality}
A recurring topic was the importance of understanding the clinical context in data collection and model development. Healthcare data collection is inherently biased by clinical decision-making processes – MRI scans are typically only performed for suspected cases, and lab tests are ordered based on clinical necessity. This selection bias poses unique challenges for developing robust multimodal models. The group stressed that understanding these clinical sampling biases is crucial for building trustworthy models. Additionally, the quality of data varies significantly across modalities, with some data types being more standardized (lab values) than others (clinical notes).

\subsubsection{Model Architecture and Specialization}
The discussion revealed insights about model architecture choices and the degree of specialization needed. While general-purpose foundation models can provide quick baselines, they often lack the fine-grained understanding necessary for complex medical tasks. The group noted that even state-of-the-art general models might struggle to outperform simpler, specialized models on specific clinical prediction tasks. This observation led to a consensus that healthcare-specific architectures, particularly those designed to handle multiple modalities effectively, may be necessary for optimal performance.

\subsubsection{Temporal Considerations and Infrastructure}
The handling of temporal data emerged as a critical challenge in multimodal healthcare AI. Healthcare systems generate complex time series data across different modalities, often with obfuscated timestamps and events not recorded in chronological order. The group emphasized the need for robust methods to handle temporal alignment across modalities and the importance of maintaining temporal coherence in model predictions. This challenge is compounded by current healthcare IT infrastructure limitations, which often make it difficult to align and synchronize data from different sources properly.

\subsubsection{Privacy and Security}
The integration of multiple modalities raises significant privacy and security concerns. Each additional modality introduces new potential vulnerabilities and increases the complexity of protecting sensitive patient information. Resource constraints were identified as a major barrier to building specialized healthcare foundation models from scratch. Many participants advocated for a pragmatic approach of fine-tuning pre-trained general models on healthcare-specific tasks, balancing computational costs with performance requirements.

\subsubsection{Future Directions}
The discussion highlighted three critical areas requiring attention for advancing multimodal foundation models in healthcare. First, there is a pressing need for robust methods to handle incomplete multimodal data and temporal alignment, particularly in real-world clinical settings where data streams are often misaligned or partially missing. Second, the development of healthcare-specific benchmarks that reflect real-world clinical conditions was identified as crucial for meaningful evaluation of these models. Third, participants emphasized the importance of improving data infrastructure to better support multimodal data integration while maintaining privacy and security standards. These directions reflect the practical challenges encountered in current implementations and represent concrete areas where focused research efforts could yield significant improvements.
\subsection{Causality}

\paragraph{Subtopic:} Many sub-areas in machine learning (e.g., Computer Vision and Natural Language Processing) have greatly benefited from having standardized benchmarks; if causality were to create such a benchmark, what should it look like? And what might be some challenges? Predictive models are widely used in many healthcare applications; how could causality be used to better develop predictive models?

\paragraph{Chairs:}
\textit{Rahul Krishnan, Johnny Xi, Trenton Chang, and Winston Chen}

\paragraph{Background:} 
Causal inference is widely applied in healthcare to derive reliable medical insights and develop robust AI tools to enhance the quality and efficiency of care. We started our discussion by having participants share their expertise in causal inference and what healthcare problems they have tried to solve using causal inference.

Standardized benchmarks have been proven effective in moving subareas of machine learning forward. We further discussed the potential of creating standardized benchmarks for causal inference in healthcare. Participants shared their thoughts on what those benchmarks would look like and what challenges the community must overcome. 

Finally, machine learning has been widely used to develop predictive models in healthcare. We further discussed how causality methods can improve predictive models. What are some considerations when building causally motivated models? 

\paragraph{Discussion:} 

We began by understanding participants' experience with causality. Most participants are familiar with the concept of causal inference, and about half of the participants have hands-on experience working with causal inference methods to estimate the effects of interest.

During the discussion, participants also shared applications in healthcare for which they have used causal methods. Most participants leverage causal inference to understand the heterogeneous effect of specific clinical treatments (e.g., the effect of chemotherapy on lung cancer patients with different genetic mutations). 
Some participants build causal reward models to better optimize for reinforcement learning agents to provide drug dosage recommendations. 

Other applications include incorporating causality into the learning process involving healthcare data and emulating clinical trials with causal models to assess the benefits of drug re-purposing efficiently. 

\subsubsection{Benchmarks and Evaluation}

Creating standard benchmarks has immensely benefited many sub-fields of machine learning. However, such a standard benchmark does not yet exist in causality and healthcare. One of the biggest challenges in creating a standard causality benchmark is that evaluating the estimated causal relationship is not straightforward. This is because ground-truth causal effects are rarely observed. 

Some participants shared their thoughts on what a principled evaluation protocol for causal inference might look like. One possibility is to leverage both observational and interventional (randomized control trial) data to assess the performance of different methods in inferring causal relationships. In this approach, causal effects may be estimated from observational data and subsequently validated on interventional data. 

Another approach for evaluating the causality method is considering the downstream application for which the causal model is built. For example, when causal inference is developed for decision-making, a meaningful approach to evaluate different methods is quantifying the decision-making performance following the effects estimated by each method. 

Interventional data can be used to accurately estimate decision-making performance by leveraging off-policy evaluation (OPE) methods. Other participants also proposed creating benchmarks by selecting applications with which true causal effects can be accurately measured. For example, in many biological and molecular applications, outcomes conditioned on different treatments can be obtained with careful experimental designs. In such cases, researchers can obtain the true causal relationship and use it to validate the performance of different causality methods.
With the recent rise of large language models (LLMs), simulation-based evaluation has gained interest in causality. Instead of estimating the performance of causal methods via OPE methods or selecting applications for which true causal relationships can be experimentally validated, creating high-fidelity simulation can be a good and cheap starting point for evaluating estimated causal insights.

This idea of simulation-based evaluation is common in social sciences such as psychology. In this field, LLMs simulate different human agents and use the simulation results to rank hypotheses and prioritize experimental validation. 
These ideas can inspire the evaluation of causality. 

Some participants also pointed out the importance of having a standardized reporting guideline. Future guidelines should consider reporting the sensitivity of estimation instead of simply reporting the estimated effects. For example, some important information to report include whether certain subpopulations disproportionately contributing to the estimated effect, and how much the estimated effect vary when modifying specific individuals in the cohort.

\subsubsection{Causality and Prediction}

Lastly, our round table discussed how causality could be used to improve predictive models in healthcare. It is important to note that although causality has many desirable properties, it is not always applicable. This is because, in practice, we often observe a trade-off between robustness and performance when building causally-motivated models. Fully causal models are often robust yet suboptimal in terms of performance. Good use of causality should consider the need for specific applications. For example, if certain applications are concerned with having the most performative model instead of leveraging causal relationships, using causal methods might not be necessary. In practice, high-performance predictive models often need to leverage non-causal correlations.

Disentangling causal and correlational representations in the learned predictive model can be a powerful future research direction. Learning such disentangled representations allows practitioners to reuse the causal components across different environments (e.g., hospitals). Then, by augmenting the causal components with environment-specifically learned correlational representations, predictive models' performance can be further enhanced to obtain the best of both the causal and correlational worlds.

Lastly, our participants shared their interests in applying causality of building foundation models in healthcare. Currently, the majority of the foundation models are built purely by leveraging correlations. It is unclear how causality can be foundation models and what some benefits of having causally-motivated foundation models are. This largely unexplored area makes a very intriguing and an impactful future research direction. 
\subsection{Challenges of Interdisciplinary Research}

\paragraph{Subtopic:} Navigating the complexities of interdisciplinary collaboration in health AI poses significant challenges, such as integrating diverse expertise and overcoming generational technology gaps. How can we effectively overcome these barriers to enhance data accessibility and collaborative efforts?

\paragraph{Chairs:}
\textit{Dennis Shung, Vasiliki (Vicky) Bikia, and Nicole Zhang}

\paragraph{Background:}
Combining expertise from computer science and clinical fields to effectively leverage health data is fraught with challenges \citep{kelly2019key}. Accessing comprehensive datasets, such as digital mammography data, often highlights barriers due to the need for clinician partnerships and extensive data curation. The process often involves significant time and effort to link or anonymize data, making it a lengthy endeavor to obtain usable datasets. Additionally, collaboration across disciplines often encounters communication barriers and technological reluctance. Engineers unfamiliar with medical intricacies and clinicians unaccustomed to machine learning paradigms often struggle to find a common language, while generational gaps in technology adoption can further complicate collaboration. Integrating and generalizing AI models within clinical workflows remains a significant challenge due to the diversity of data modalities and the complexity of healthcare settings. The need for model-agnostic approaches is critical to ensure that AI tools are adaptable and effective across different healthcare environments.

\paragraph{Discussion:}
Addressing these interdisciplinary challenges requires strategic partnerships and dedicated educational initiatives. The formation of robust collaborations with clinicians and the investment in continuous interdisciplinary education are essential to bridge knowledge gaps. 

Engaging in direct conversations with clinicians not only helps computer scientists understand the intricacies of medical practice but also allows clinicians to grasp the potential and limitations of AI technologies. 
Additionally, proactive measures such as attending interdisciplinary conferences provide valuable opportunities for networking and learning from leading experts in both fields \citep{patel2024crucial}. Conferences often feature workshops and symposiums that specifically focus on integrating AI into healthcare, offering practical insights and fostering collaborations. Similarly, reading interdisciplinary journals and books can enhance understanding. Exposure to each other's work environments can further enhance collaboration. Visits to clinical settings can help computer scientists see firsthand the challenges clinicians face, while visits to tech labs can show clinicians the complexities of developing and training AI models. This mutual exposure not only deepens understanding but also builds trust and respect between the disciplines, paving the way for more effective and empathetic collaborations.

Assembling interdisciplinary teams that include members proficient in both health and computer science is crucial for advancing AI integration into healthcare. These teams benefit immensely from having \textit{bilinguals} members who can translate between the technical language of computer science and the clinical language of healthcare. These bilinguals make complex concepts accessible and actionable, serving as vital bridges within the team.

Additionally, emerging systems such as virtual labs and AI agents can potentially enhance efficiency and productivity in these collaborations \citep{swanson2024virtual}. Virtual labs provide advanced simulation platforms for virtual meetings and collaborative sessions, mimicking real-world interactions and experiments. LLM agents are designed to simulate real meeting scenarios by undertaking specific roles within a team, such as note-taking, summarizing discussions, generating code, or even proposing solutions based on the dialogue. These agents not only automate routine tasks but also provide models of communication and problem-solving that human team members can learn from.

Furthermore, developing adaptive and inclusive communication strategies that cater to varying expertise levels is essential \citep{sankaran2021practical}. Coupled with the adoption of modern collaborative tools, these strategies help bridge the significant gaps between different generational and disciplinary perspectives. This approach ensures that all team members, regardless of their primary discipline or familiarity with new technologies, can fully engage and contribute to the project’s success.

Integrating AI into healthcare through interdisciplinary research presents formidable challenges but also opens significant opportunities for innovation. By fostering understanding and cooperation between fields, leveraging AI tools for enhanced collaboration, and addressing the persistent issues of data access and usability, we can pave the way for more effective and transformative healthcare innovation.

\subsection{Health Economics, Policy, and Reimbursement}

\paragraph{Subtopic:} 
What is the appropriate role for AI/ML powered health economic models in decision making? How do we enable and encourage these positive applications?

\paragraph{Chairs:} \textit{Ian Cromwell, Shuvom Sadhuka, and Ross Duncan}

\paragraph{Background:}
“Health Economics” is a discipline that applies economic theory to understanding the production and consumption of health and healthcare. Health Economics most prominently focuses on 1) determining value (or comparisons of value: efficiency, effectiveness) of healthcare goods and services consumed/provided, 2) how to optimize allocation of scarce resources across patients, and 3) understanding the incentives and expected behavior of various actors within the market for health. The minimum cast includes patients (consumers), providers (producers), and payers (often insurance and so distinct from consumer). It is distinct from standard economics as “health” is not appropriately modelled as a typical consumer good and the market for health structurally differs from requirements for standard economic theory.  For example, demand for life-saving care is inelastic – that is, the patient-consumer is willing to pay whatever it takes to receive care. Health is also not a good that can be transferred or purchased directly, it requires services delivered by experts with knowledge that consumers are often not able to independently assess. Finally, due in part to the large upfront costs required for intensive care episodes, the payer is not always – or even often – the same as the consumer of care. These, among other structural factors, form a special case that “Health Economics” studies – often with an eye to understanding how to optimally allocate scarce resources. Contemporary understanding of “health” is of a kind of personal capital, as outlined in Grossman \citep{grossman} How do we determine which healthcare services and technologies are best suited to deliver the most benefit for resources required \citep{cadth}?

\paragraph{Discussion:} Many participants did not have any direct health economics experience or background, and so a basic outline of health economics and its role was provided by the senior chair. Key areas of distinction where to explain the role of “quality-adjusted life years” (QALY) and “disability-adjusted life years” (DALY) often used as an outcome measure in health economic work. These are both psychometric instruments that try to capture both the length of lifetime gained by an intervention, and the quality (or capabilities) of that time \citep{Howren2013}. Methodological debates still occur over those instruments, but health economics will also use direct outcome measures and derived utility to measure benefits. Secondly, that health economics is an aid to decision-making at the policy and population level – it does not focus on specific, individual instances of care. That is, a health economist may assess whether it is worth deploying AI-assistants in hospitals, but not whether the assistant should be used in any specific instance of care.

\subsubsection{Where do health economic standards come from?}

Standards for health economic evaluation tend to be set by agencies tasked with assessing new healthcare technologies, drugs, and other interventions. The clearest example of this is the United Kindgom’s National Institute for Health and Care Excellence (NICE), an organization which determines which (typically novel) healthcare drugs or technologies should be provided by the National Health Service, and also wrote the manual on economic evaluation \citep{national2022nice}. A similar Canadian organization with a more advisory role is Canada’s Drug Agency (CDA-AMC, formerly CADTH), which has and continues to produce guidelines for health technology assessment and health economic evaluation \citep{cadth}. CDA-AMC led the adoption of open source practices in health economics, which has been adopted as the Canadian standard. The centrality of these organizations in determining standard practice is of consequence when determining how to best deploy ML/AI in health economics but are perhaps not ideally situated to lead by example where more regulatory intervention may be necessary. Certainly there is a need to standardize which models are used and why, as competing practices continue to proliferate. Vithlani et al’s 2023 systematic review of best practices in conduct and reporting may be useful for shaping future discussion \citep{vithlani2023economic}.

\subsubsection{What can AI/ML do for Health Economics?}

From a decision making perspective, private for-profit insurers in the United States have already adopted models trained to determine when to approve or deny care. Such models can be tuned to hit targeted rates and maximize billing. This does not constitute health economics as typically practiced – as this involves automation of individual care decisions, but the question of whether – from a societal perspective – the cost savings from denying “unnecessary” care outweigh the instances in which care is inappropriately denied. When such errors inevitably do occur, it is unclear who would be responsible and held accountable: the model creators? The provider? The insurers who deployed the model? 
Discussion then turned to less controversial applications of AI/ML in healthcare settings, such as using LLMs to help write authorization letters and similar documentation to expedite administrative burden. Sufficiently advanced LLM models may also be able to assist patients with health concerns – as patients are already engaging with tools like ChatGPT to learn about their condition. This may save time and money associated with concierge physician visits. 
In the literature, the International Society for Pharmacoeconomics and Outcomes Research (ISPOR) has produced the PALISADE Checklist of ML methods in Health Economics. The review suggests that ML could enhance 1) specificity of cohort selection, 2) identification of independent predictors and covariates of health outcomes, 3) predictive analytics of health outcomes, 4) causal inference, and 5) reduction of structural, parameter, and sampling uncertainty \citep{padula2022machine}. 

\subsubsection{What regulation is needed to guide AI/ML in Health Economics?}

The example of AI in health insurance prompted discussion about what sort of oversight may be needed, and what functions were appropriate for such models. Some participants raised concerns about accountability for decisions made by models, but others felt that the limitations of AI must be considered in comparison to the available alternatives. Physicians themselves are capable or error, and if it is the difference between no care and AI-moderated care then the latter would still presumably be preferable. Regulation could also stymie innovation and so some suggested that a “Consumer Reports” type transparency around AI/ML models may be preferable. 
The suggestion that AI/ML models for use in healthcare should go through the Food and Drug Association (FDA) also arose – those software is not typically held to those standards the unique role of health may urge particular caution before bringing to market. 
There was an emerging consensus that AI/Models deployed in healthcare decision making should be monitored, and that the monitoring should have regulatory enforcement. However, how to ensure scalable oversight is still an open question.

\subsection{Integrating AI into Clinical Workflows}

\paragraph{Subtopic:} What level of evidence is required to decide to deploy or retire a model? How do emerging governance structures for AI deployment in hospitals (e.g., Chief AI officer) help these efforts? 

\paragraph{Chairs:} \textit{Adarsh Subbaswamy, Ayush Noori, and Elizabeth Healey}

\paragraph{Background:}
AI has the potential to transform clinical medicine through improved risk prediction, triaging, decision support systems, and helpful administrative tools. Integration of AI models and systems into clinical workflows may improve both outcomes and efficiency in medicine. However, there are many risks and considerations at play when deploying AI into medical systems that must be addressed. For example, AI models must be monitored to determine when they should be deployed and when they should be retired, and few standardized protocols exist for how to manage and monitor these systems over their lifetimes. Furthermore, some models may be more prone to distribution shift, privacy breaches, biases against protected classes, or other risks – therefore, the type of oversight necessary depends on characteristics of individual models. One approach to mitigate these harms include specialized governance structures like the FDA or Chief AI Officers.

This roundtable discussion spanned a breadth of considerations for deployment of AI into clinical workflows, including model considerations, regulatory structures, and practical limitations. The discussion was purposefully designed to bring together unique perspectives from a variety of stakeholders.  

\paragraph{Discussion:}

There was great interest in this topic at ML4H, with over 30 attendees at this roundtable on integrating AI into clinical workflows. Many participants from various backgrounds, including clinicians, industry professionals, and academics, contributed to the discussion. Below, we summarize the discussion by theme. 
 
\subsubsection{Gap Between Research and Deployment}
The conversation started with a discussion on what it takes to bring AI methods to clinical deployment. Many participants pointed out gaps that exist that prevent promising research from making it to a clinical setting. These gaps often center around ways that models are evaluated. Hospital integration of AI models relies on demonstrated clinical utility – a metric that is often measured during research and development of models. Moreover, for certain tools, such as those that include LLMs, there may be disagreement between providers about specific medical decisions.  The variability in human annotation further complicates the evaluation process of certain systems.  How can we measure the performance of decision support tools when human annotators do not agree?
 
The conversation also touched on other deployment considerations, such as distribution shifts.  The effect of distribution shifts should be quantified and studied during the development of models. It also may be prudent to set standards for certain types of deployments for required routine testing to calibrate models. 
Participants also discussed operational challenges with deploying systems in medical systems and cited the importance of including the end-user during the development of systems. Many new tools will require training from medical staff to properly utilize them. Human-computer interaction research on deployment of clinical AI will inevitably be essential to understanding the impact of systems in the clinical workflow, and for understanding how to continually improve these systems based on feedback. 

\subsubsection{Incentives}
The discussion briefly touched on incentives for both researchers and hospitals in this space. In particular, concerns were raised about how to make technology more equitable. Some hospital systems may be better equipped with infrastructure for integration of these systems. When developing AI systems, available infrastructure should be considered such that feasible deployment is not only limited to high-resource hospital institutions.
 
\subsubsection{Monitoring and Surveillance}
Many individuals highlighted the importance of regulatory structures and monitoring protocols specifically for AI deployments. There was agreement that different AI tools require different regulatory processes, and deciding how these tools should be monitored and who should regulate them is an area of importance.  The FDA has been involved in the regulation of biologics and devices. Clinical AI can take a variety of forms with a wide range of safety risks. It is not obvious now what AI models should be defined as medical devices. For high-risk systems, the idea of monthly audit reports was discussed. Most individuals agreed that the type of surveillance will depend on the type of model and specific risk of distribution shifts.

\subsubsection{Looking Ahead Towards Deployable Clinical AI}
Among the diverse group of individuals at the roundtable, many cited the importance of interdisciplinary collaborations including both AI experts and clinicians. The involvement of clinicians in the early stages of projects is paramount to developing methods with the potential to impact clinical care. The group discussed the importance of individuals who work at the intersection of AI and medicine who can bridge the gap to deployment.

\subsection{Public Datasets and Benchmarks}
\paragraph{Chairs:} \textit{Rahmat Beheshti, John Wu, and Adibvafa Fallahpour}
\paragraph{Background:}

The healthcare AI community faces significant challenges in establishing standardized benchmarks and utilizing public datasets effectively. Despite the availability of several large-scale medical datasets like MIMIC-CXR and MIMIC-IV, researchers encounter substantial difficulties in dataset preprocessing, standardization, and reproducibility of results.
\paragraph{Discussion:}
The roundtable discussion revealed several critical challenges and opportunities in healthcare AI benchmarking, which can be organized into four main themes:

\subsubsection{Dataset Standardization Challenges}
A significant concern raised was the heterogeneity in dataset usage, particularly evident in studies utilizing MIMIC-CXR. Participants highlighted the unbelievable heterogeneity in how researchers split the data, with no clear consensus on task definitions or evaluation metrics. While image-based models often come with standardized preprocessor classes, there is no equivalent standard for EHR data processing. This variance extends beyond simple data splitting to encompass the entire preprocessing pipeline. Of particular concern was the clinical validation of preprocessed variables, where participants noted a significant disconnect between dataset variables and clinical understanding, with estimates suggesting that up to half of the variables might be incorrectly defined or interpreted. The variability extends to fundamental choices such as whether to use radiological findings or impressions and how to define train versus test distributions. The preprocessing pipeline has effectively become an integral part of model design and performance evaluation. The emergence of the MEDS data format represents a promising attempt to standardize EHR data representation and preprocessing, though adoption remains a challenge due to limited incentives for standardization.

\subsubsection{Benchmark Development and Adoption}
The fundamental question of benchmark definition emerged as a critical issue. Before implementing any benchmark system, participants emphasized the need to clearly define the benchmark's objectives and purpose. Current adoption rates of existing benchmarks remain concerningly low, with longitudinal EHR data benchmarks being used by a maximum of nine studies for comparison purposes. The discussion revealed that many researchers focus on developing increasingly complex models rather than conducting comprehensive baseline evaluations. A significant technical challenge emerged regarding the dependency of models on specific preprocessing pipelines, making it difficult to compare models across different preprocessing approaches. Models are often essentially hardcoded for specific types of preprocessing, such as particular sampling rates, making standardization particularly challenging.

\subsubsection{Dataset Availability and Utilization}

The discussion challenged the common perception of limited dataset availability in healthcare AI. Participants noted the existence of substantial public data resources, including approximately 700,000 public ICU patient records and emergency datasets from Iran. However, the real challenge lies not in data availability but in utilization. Researchers face significant barriers including time constraints for data exploration and processing, and limited awareness of available datasets. Even when datasets are publicly available, access can be complicated by payment requirements, non-responsive data holders, and complex licensing agreements. These practical barriers often prevent researchers from exploring alternatives to commonly used datasets like MIMIC. Some groups have successfully merged multiple public ICU datasets, demonstrating that meaningful generalization becomes possible at the scale of seven combined datasets.

\subsubsection{Future Directions and Recommendations}

The path forward requires addressing multiple challenges simultaneously. Participants advocated for the development of multi-population benchmarks that span different healthcare settings, from ICU to standard population care, and diverse geographical locations. The potential of LLMs for dataset harmonization was discussed as a promising avenue for future research. The group emphasized the importance of implementing standardized sub-splits specifically designed to evaluate bias and fairness across different patient populations. Collaborative discussions about cohort selection and feature definition should be conducted openly, potentially through platforms like GitHub, rather than being buried within academic papers.

High-quality benchmark papers should provide comprehensive documentation of the complete pipeline from raw data to predictive modeling, accompanied by well-documented code. The inclusion of completely open demonstration datasets free from licensing restrictions was identified as crucial for enabling code validation and pipeline reproducibility. Papers should report inter-annotator agreement for labels and provide detailed discussions of cohort selection methodology. Participants emphasized the importance of reproducibility and transparency in all aspects of the benchmarking process, including detailed metadata documentation, feasible range specifications, and feature descriptions.

Several critical questions emerged that warrant further investigation. The community must determine the optimal balance between comprehensive feature sets and expert-selected features in ICU datasets. The challenge of incentivizing researchers to adopt updated datasets and reproduce baseline results remains unresolved. The potential role of meta-reviews in standardizing benchmark practices requires further exploration, as does the development of better frameworks for supporting cohort selection discussions in the academic community.

\subsection{Health AI in Low- and Middle-income Countries}

\paragraph{Subtopic:} What are the unique challenges and opportunities of AI in low- and middle-income countries? How can we help these countries prepare and prevent epidemics and pandemics? How can AI be adapted to meet such challenges?\citep{mollura2020artificial}

\paragraph{Chairs:} \textit{Megan Coffee, Wenqian Ye, and Brighton Nuwagira}

\paragraph{Background:} 

Developing and deploying AI systems in low- and middle-income countries presents unique challenges and opportunities. These countries often face limitations in healthcare infrastructure, fragmentation in health systems and data management, and obstacles in accessing healthcare. These hurdles require innovative and context-specific AI solutions. For instance, AI has demonstrated promise in addressing infectious diseases such as COVID-19~\citep{jiang2020towards}, Mpox~\citep{cao2024mpoxvlm}, and Tuberculosis~\citep{ahmed2023topo} by improving diagnostics, clinical decision support, forecasting outbreaks, and optimizing resource management. However, using small or biased datasets in these settings increases the risk of spurious correlations~\citep{ye2024spurious}, undermining model robustness and fairness. In addition, data collection efforts in these regions often encounter significant obstacles, as standardized electronic health records are often not available. There are other infrastructural limitations, lack of standardization, and the need for effective collaboration between clinics and across geographies. International organizations like the World Health Organization (WHO), and non-profits like the International Rescue Committee (IRC), as well as Ti Kay, which has worked in Haiti, play a crucial role in facilitating these partnerships and ensuring ethical, equitable, and effective use of AI, in addition to direct partnerships with local universities and clinics. Building trust in AI systems among healthcare practitioners requires focusing on interpretability, cultural relevance, and practical utility. Engaging with local stakeholders and tailoring solutions to specific healthcare challenges are critical steps to ensure the successful integration of AI into clinical workflows in these settings.



\paragraph{Discussion:}
The discussion began with an exploration of the challenges and opportunities in leveraging Health AI for low- and middle-income countries, focusing on the interplay between data availability, infrastructure constraints, and the potential for innovative solutions. A key question emerged: \textit{Should Health AI in these regions prioritize a data-centric or algorithm-centric approach, particularly given the limited access to high-quality data?} 
The discussions highlighted several opportunities and strategies to overcome these challenges.

\subsubsection{Synthetic Data}
Synthetic data has been recognized as a transformative tool for mitigating data scarcity in low- and middle-income countries. These datasets, generated using advanced generative AI techniques, such as Stable Diffusion~\citep{rombach2022high}, simulate realistic and anonymized data that maintain critical statistical and clinical properties. Additionally, efforts are underway to leverage 3D generative models to create synthetic skin textures, 3D anatomical models, and digitally rendered medical images~\citep{kim2024s}. They offer a means to train models effectively where real-world data is scarce, while also addressing privacy and ethical concerns in data-sharing practices. However, participants underscored the necessity of implementing rigorous validation protocols to ensure that synthetic data aligns with the specific healthcare needs and contexts of these regions, while also accounting for the limitations in the generalizability of these models. It is important that models for low- and middle-income countries are not reliant on synthetic data as gap fillers, but instead substantive effort is made to collect representative data.

\subsubsection{Spurious Correlations, Model Robustness, and Generalization}

Participants highlighted the critical challenge posed by spurious correlations, which arise when models rely on superficial patterns in training data that fail to generalize to new or diverse distributions. This issue is particularly pronounced when AI models developed in high-resource settings are deployed in low-resource environments, where differences in demographics, healthcare practices, and disease prevalence often render such models ineffective. To overcome these limitations, the discussion emphasized the need for diverse and representative datasets that capture these variations. Additionally, employing techniques such as domain adaptation and transfer learning was identified as essential to enhance model robustness and ensure reliable performance across heterogeneous contexts.

\subsubsection{Stakeholder Engagement}
The importance of engaging all relevant stakeholders in data collection and model deployment processes was a recurring theme. Collaboration with local healthcare professionals, policymakers and community leaders ensures that AI solutions are contextually relevant, practically feasible, and aligned with local needs. Co-developing AI systems with local stakeholders foster trust and accelerates adoption in clinical workflows. These partnerships also promote long-term sustainability by integrating AI into healthcare systems in a meaningful way.

\subsubsection{Partnerships Between Developed and Low-Resource Regions}
Robust partnerships between researchers in developed countries and those in low- and middle-income regions were recognized as vital. Such collaborations facilitate knowledge transfer, financial investment, and infrastructure support, enabling the development of state-of-the-art AI models that address the specific challenges of low-resource settings. Importantly, these partnerships are mutually beneficial, driving innovation by exposing researchers to diverse healthcare contexts and fostering solutions that address the needs of both high- and low-resource regions.

\subsubsection{Data-Centric vs. Algorithm-Centric Approaches}
The discussion weighed the relative importance of data-centric versus algorithm-centric approaches. While sophisticated algorithms are crucial, there was consensus that improving the quality, representativeness, and availability of data should be the priority in low-resource settings. Without adequate and reliable training data, even the most advanced algorithms are unlikely to perform effectively.

\subsubsection{Ethics and Trustworthiness}
Building trust in AI systems was identified as a critical priority. This requires ensuring that models are interpretable, transparent, and ethically sound. Efforts by international organizations like the WHO and IRC, along with collaborations with other smaller non-profits organizations and local academics and clinicians, play an essential role in promoting ethical AI adoption. Participants also noted the importance of addressing biases and spurious correlations to enhance the fairness and reliability of AI solutions.

The discussion concluded that advancing Health AI in low- and middle-income countries demands a holistic and collaborative approach. Combining data collection, stakeholder engagement, international partnerships, and robust model development can create impactful and equitable AI solutions tailored to these regions' unique needs. By prioritizing data quality and fostering trust, the global health AI community can make meaningful strides toward improving healthcare outcomes in these settings.





\subsection{Bias and Fairness}

\paragraph{Subtopic:} What are the best practices for continuous evaluation of deployed models, given the rise in the popularity of foundation models? Given that resources and expertise required for the requisite local recalibration before model deployment and continuous evaluation of the model one deployed are not available to most health systems, how can we realistically ascertain \textit{first do no harm}? Given how higher education is notoriously slow to change, how do we achieve the AI literacy needed for fail-safe human-AI systems?

\paragraph{Chairs:‌}
\textit{Leo Anthony Celi, Alina Peluso, and Amin Adibi}

\paragraph{Background:}
Philosophers of science have long argued that scientific objectivity and the value-free ideal are myths, particularly in how studies are conceptualized, funded, interpreted, published, and publicized \citep{reiss_scientific_2020}. Even the production of scientific evidence involves a "garden of forking paths" of countless decisions with equally valid alternatives \citep{gelman_garden_2013}, as exemplified by models used to guide the COVID-19 response \citep{harvard_value_2021}. This understanding suggests that it matters who does the research \citep{charpignon_does_2024}. As such, epistemic humility, plurality, and incorporation of diverse viewpoints and lived experiences should be at the heart of the fairness movement in AI.

Recent advances in foundation models have aligned economic and political forces to push for rapid AI deployment, even in traditionally risk-averse fields such as medicine. Although some large health systems in the US have created roles such as Chief Health Data Officer \citep{beecy_chief_2024}, most hospitals in the developed and developing world lack the expertise and resources required for proper AI governance, deployment, and continuous evaluation. A complete overhaul of medical education, clinical validation, regulation, and biomedical innovation appears necessary to equip health systems with the expertise and infrastructure to innovate and improve care, while safeguarding against the risk of harm, particularly to the most vulnerable populations. 

\paragraph{Discussion:}
The roundtable discussion was catalyzed by a recent incident of inappropriate stereotyping in research. Society often emphasizes characteristics such as gender, race and ethnicity in descriptions, even when these factors may not be essential to the discussion. This type of stereotyping reduces the complexity of individuals or groups to simplistic labels. While these traits can be relevant in some contexts, such as discussions of systemic inequality, it is important to assess whether their inclusion is necessary and respectful. One way to avoid such pitfalls for us, the researchers, is by actively seeking out diverse team members who are willing to challenge our assumptions and call out our biases. 

\subsubsection{Embracing Epistemic Humility and Plurality}
Participants noted the speed at which the field is progressing, making it almost impossible for any single person to be an expert in all aspects of AI research. Incorporating social scientists—particularly those with expertise in feminism, race theory, and ethics— and patient views into our teams enriches AI research and ensures that marginalized voices are heard. This diversity of expertise helps shape AI systems that are not only more inclusive but also more reflective of the complexity and nuances of human experience. However, despite the critical importance of interdisciplinary perspectives in AI development, even leading universities may be reluctant to recruit multidisciplinary scientists for faculty positions if they lack traditional technical expertise.

The discussion also emphasized the role of value judgments in research and the importance of incorporating diverse viewpoints from various cultural backgrounds. One participant shared how working with Indigenous researchers revealed alternative frameworks for ensuring research benefits affected communities through relevance, equity of partnership, and data sovereignty. Epistemic humility and plurality require engaging with others in a way that values their ways of knowing, unique identities, and perspectives, while remaining mindful of the social and historical contexts that shape our own views.

\subsubsection{Epistemic Justice and Algorithmic Humility}
Epistemic injustice and oppression constrain scientific progress by excluding vital perspectives and insights \citep{dotson_conceptualizing_2014, fricker_epistemic_2007, kay_epistemic_2024}. When entire groups face marginalization from knowledge production and scientific discourse, we lose insights that could challenge prevailing assumptions and enrich our understanding. To paraphrase an old adage: "If everyone is thinking alike, then no one is really thinking." This is where algorithmic humility — AI designed to offset human arrogance and ignorance - comes in. Algorithmic humility is an approach to AI system design that explicitly acknowledges the boundaries of both human and machine intelligence. Unlike conventional AI approaches, systems built on algorithmic humility principles incorporate mechanisms to actively counteract human overconfidence and uncritical acceptance of automated outputs \citep{kompa_second_2021}. Through this framework, algorithmic humility seeks to develop AI systems that not only address human cognitive biases but also resist becoming unquestioned sources of authority themselves.

\subsubsection{Structural Barriers to Inclusive Research}
Participants identified several systemic barriers to achieving truly inclusive ML research in healthcare. The \textit{publish or perish} culture continues to incentivize quantity over quality and methodology over impact. Conference accessibility remains severely limited by visa restrictions — with wait times for US visa interviews exceeding two years in some countries — and prohibitive travel costs \citep{hutson_canada_2018, owusu-gyamfi_exhausted_2024, thompson_canadian_2024}. Even when researchers from underrepresented groups participate, power dynamics can make it challenging for junior scholars to speak up about bias and discrimination without risking career advancement. The discussion also highlighted how major conferences like NeurIPS, ICML, and ICLR often prioritize the novelty of methodology and larger datasets over real-world impact, perpetuating existing power structures. Extractive AI practices exacerbate barriers to inclusivity by exploiting the resources and labor of the global majority without contributing to the societies that bear the costs of AI development \citep{crawford_atlas_2021, li_making_2023, luccioni_power_2024}.

\subsubsection{Novel Approaches to Measuring Disparities}
While much work on bias and fairness has focused on technical approaches, apparent technical solutions can mask deeper systemic issues. Participants shared examples of how such solutions can fall short. For example, attempting to debias models for multiple protected attributes simultaneously can degrade the model’s performance to the point of clinical irrelevance, achieving equality through universal poor performance. 

Ensuring the safe and effective deployment of AI requires continuous evaluation, transparency, and human oversight, with rigorous testing and active involvement of clinicians in decision-making \citep{adibi2020, Blumenthal2024}. Strong regulation and ongoing model evaluation are key to ensuring AI tools are ethically integrated. Fairness in AI is not about achieving perfect equality but preventing harm to vulnerable populations. 

Participants highlighted innovative methodologies for identifying and measuring healthcare disparities without relying on traditional demographic categories. One innovative approach introduced the concept of \textit{care phenotypes}, patterns in care delivery that reveal systematically underserved patient populations. For instance, research demonstrated that the heaviest intubated patients were not only turned in their beds less frequently than others, but also, inexcusably, received mouth care less often \citep{weishaupt_care_2025}. This approach revealed systematic disparities in care delivery that map to social determinants and potentially accounted for intersectionality, without requiring explicit collection of demographic data. Another example showed how language barriers affected the frequency of blood sugar testing, as staff avoided waking non-English-speaking patients during night shifts when interpreters were not readily available.

\subsubsection{Education and Training Reform}
Healthcare systems often lack the expertise and infrastructure for responsible AI deployment. The roundtable highlighted the need to fundamentally reimagine medical and technical education, with concerning findings that patients sometimes perceive chatbots as more empathetic than human doctors \citep{ayers_comparing_2023}. Participants emphasized the need to reimagine medical education to nurture empathy and incorporate systems thinking and critical analysis of technology. This includes teaching students to question assumptions, understand local context, and consider long-term impacts rather than just immediate clinical outcomes. A systems-level approach is needed to integrate epistemic humility into AI and medical curricula. Empathy and interdisciplinary collaboration should be prioritized to equip students with the tools to address the ethical complexities of AI and its intersection with healthcare. 

\subsubsection{Optimism and Call for Action}
The discussion yielded several concrete recommendations, including expanding conference accessibility through virtual participation options and events in diverse locations and languages, advocating for diverse representation in research teams and paper authorship, developing new frameworks for clinical validation that account for local context and temporal shifts in care patterns, creating dedicated courses on epistemic humility and plurality in computer science, reforming publication incentives to value impact and ethical considerations over methodological novelty, and building research teams that include social scientists and bioethicists and incorporate patient perspectives from project inception.

While acknowledging the magnitude of needed changes, participants expressed optimism about growing awareness and willingness to confront bias in the field. They emphasized that progress requires moving these conversations from the margins to the center, with everyone taking responsibility for change within their sphere of influence. The discussion highlighted how even those early in their careers can make meaningful contributions, citing examples of PhD students choosing to focus on bias and fairness despite institutional pressure to pursue more traditional technical topics. Success will require sustained effort to reimagine education, research, and clinical validation while maintaining focus on benefits to vulnerable populations.

\subsection{Clinician-AI Interaction}

\paragraph{Subtopic:} What are the primary use cases of AI tools in clinicians’ work? What are challenges, limitations, and concerns of adopting AI tools in clinical workflows? What are the lessons learned from evaluations of Clinician-AI interactions? 

\paragraph{Chairs:}
\textit{Shannon McWeeney, Yuan Pu, and Shannon Zejiang Shen}

\paragraph{Background:} 
The development of many clinical AI tools is predominantly driven by the model creators: the AI researchers identify the underlying machine learning problem and build competent models with good accuracy in such tasks.
However, it is often an afterthought how clinicians can interact and collaborate with these powerful but imperfect AI models in practice, and the efficacy of such AI systems is often evaluated in isolation from clinical workflows ~\citep{10.1145/3290605.3300468}.
In this roundtable discussion, we examine clinician-AI interactions in real-world deployments. 
With a cross-disciplinary audience of around 10 clinicians and 20 computer science/AI researchers, we identify both gaps and opportunities for building effective human-AI collaboration systems in clinical settings.

\paragraph{Discussion:}
Several key themes emerged during the discussion. 

\subsubsection{Use Cases of AI in Clinical Workflow}

A wide range of AI support is developed across different stages of clinical practice, from streamlining operational workflows to supporting complex clinical decisions ~\citep{10.1145/3582430}. 
The participants discussed several imminent AI applications for automating routine tasks like patient check-ins, prescription management, and payment processing. 
Beyond these operational improvements, AI systems have demonstrated promising capabilities in information synthesis, for example generating patient history summaries ~\citep{keszthelyi2023patient} and explainable X-ray reports ~\citep{DEPERLIOGLU2022152}. 
Moreover, recent research shared during the discussion has explored both AI-powered diagnostic predictions ~\citep{10.1001/jama.2023.22295} and LLM-based chatbots that help interpret risk scores and facilitate access to relevant medical guidelines ~\citep{chan2023assessingusabilitygutgptsimulation, 10.1145/3613904.3642024}.

Different modes of Clinician-AI interactions were discussed: ambient, passive, and proactive. Ambient systems, such as automated scribing tools, work continuously in the background to capture and organize patient-clinician interactions, freeing healthcare providers to focus fully on patient care ~\citep{doi:10.1056/CAT.23.0404}. Passive support comes in the form of on-demand tools like chatbots, which clinicians can consult to help navigate complex medical records and surface critical information for decision-making. The third category, proactive support, represents systems that actively monitor clinical workflows and autonomously surface relevant information and recommendations to clinicians~\citep{jiang2023conceptualizing}, enhancing their work without adding cognitive burden.

\subsubsection{Challenges} 
The roundtable discussion explored both the challenges and opportunities in fostering effective clinician-AI interactions. Participants highlighted clinicians' mistrust of AI and their misunderstanding of its applications as significant barriers. They emphasized that addressing these issues will require collaboration and thoughtful design to build trust and enhance the integration of AI into clinical practice.

\paragraph{Socio-technological challenges in practical AI deployment} 
Roundtable participants frequently mentioned that clinicians often show reluctance to adopt AI predictions, even in scenarios where AI demonstrates high accuracy. 
For instance, in a study shared during the session, clinicians disregarded 20\% of AI-generated predictions, even when those predictions were 100\% accurate ~\citep{10.1001/jama.2023.22295}. 
The group suggested that this hesitancy may stem from clinicians' strong preference for relying on their professional judgment developed through extensive training and hands-on experience. 
Resistance to revising decisions based on AI recommendations was further linked to cognitive dissonance. 
As several participants observed, healthcare professionals often experience notable discomfort when faced with contradictory opinions, whether from human peers or AI systems. 
Accepting such recommendations may be perceived as admitting an error, creating tension with a clinician's professional self-image and confidence. 
This psychological barrier can lead to an unconscious dismissal of the AI's inputs, even when the evidence strongly supports them.

\paragraph{Trustworthiness of the model outputs} 

Concerns about accuracy, accountability, and ethics also undermine trust ~\citep{HENGSTLER2016105}. 
Clinicians frequently perceive AI systems as ``black boxes,'' making it difficult to understand and trust their recommendations. 
Ethical issues, such as potential biases in AI models and the privacy of patient data, further complicate this trust dynamic.

For example, systems that use heatmaps (or activation maps) to explain model predictions for medical imaging often fail to present reasoning in an intuitive and easy-to-understand way for clinicians without AI background. 
Clinicians often develop their own interpretations of the outputs, increasing their cognitive burden and reducing confidence in the tools.
Participants also notes that it is very hard to identify the occasional errors in the AI ambient scribing system, thus making it less practically useful.
In such cases, participants emphasized that clinicians worry about who will be held responsible for decisions influenced by AI, particularly in high-stakes scenarios.

\paragraph{Disconnect from real-world case complexities} Participants also pointed out the disconnect between AI models and real-world patient dynamics. 
For example, clinicians expressed concern that real-time factors like a patient’s emotional state are often overlooked and not adequately represented in the training data used to develop AI systems ~\citep{chan2023assessingusabilitygutgptsimulation}. 
More broadly, some participants are not certain about whether synthetic datasets (used in AI deployment) can accurately reflect the variability and complexity of real-world scenarios. 
This gap limits the relevance of AI recommendations in clinical settings and contributes to skepticism. 

\subsubsection{Opportunities} 

\paragraph{Towards holistic evaluation at the model development stage} 
Going beyond only evaluating competence of the models, participants emphasized the need to capture multiple dimensions that reflect real-world user needs in clinical settings. 
For example, when evaluating AI generated X-ray reports, rather than just assessing the ``accuracy'' through lexical overlap between the generation and reference doctor's writing, one should consider metrics that directly address physicians' information needs. 
These include assessments of readability, completeness of critical findings, and the cognitive effort required to locate decision-relevant information.
Through collecting first-hand subjective feedback from clinicians users, the developers can better understand potential challenges and limitations of the AI systems in real-world settings, and iteratively refine the models to better align with clinical workflows.

\paragraph{Rethinking the mechanisms of using AI} 
Participants also suggested a reframing of how AI systems are positioned within clinical use cases. 
For example, rather than providing definitive answers, AI systems could be designed to prompt clinicians to engage in deeper analytical thinking (system 2 reasoning) -- they can prompt the clinicians to think step by step and highlight potentially overlooked patterns or suggesting alternative diagnostic considerations.
This approach aligns with how clinicians naturally collaborate with colleagues, where discussion and collective reasoning often lead to better outcomes than isolated decision-making.
The group also mentioned that incorporating AI education into medical training can help future clinicians develop a nuanced understanding of both the capabilities and limitations of AI tools.

\subsection{Drug Discovery and Development}

\paragraph{Subtopic:} 
Integrating Foundation Models: How can foundation models streamline specific stages like target identification, lead optimization, and clinical candidate selection, and what measurable impacts have been observed?  Mitigating Risks in AI Integration: What strategies address risks like data bias, overfitting, or lack of interpretability when integrating AI into drug discovery pipelines? Discovering Novel Biological Knowledge: How have ML models uncovered new biological mechanisms or pathways, and what are the key challenges in simulating complex processes like protein-ligand interactions? Enhancing Chemical Design and Synthesis: How can AI improve understanding of computational compound generation/modification and synthesis planning to accelerate the design of novel, drug-like molecules?

\paragraph{Chairs: }
\textit{Michael Craig, Zuheng Xu, and Geoffrey Woollard}

\paragraph{Background:}

AI and ML continue to rapidly advancing in drug discovery and development, accompanied by growing biological and chemical data and improved computational power. AI/ML is increasingly present in science broadly considered in terms of perspectives, language, tools, and community (both research and business). Computational biology, broadly considered, has become increasingly integrated with AI/ML communities in statistical inference, simulation, high performance computing, numerics, and data science \citep{Cranmer2020,Lavin2021,Hennig2022,Wang2024}. In recent years, there has been increased effort to integrate AI/ML methods into key stages of the drug discovery pipeline, including target identification \citep{hassanzadeh2016deeperbind}, lead compound discovery \citep{macedo2024medgan}, lead optimization \cite{pmlr-v108-korovina20a}, and candidate selection, and thus demonstrate tangible impacts. Furthermore, there is an overly-familiar claim that AI/ML perspectives will help the discovery of novel biological mechanisms and pathways, shedding light on complex interactions such as protein-ligand binding \citep{rezaei2020deep} and protein design \citep{TheNobelCommitteeforChemistry2024}.

The activity at NeurIPS 2024 showed attempts to develop foundation models (meaning large models trained on large amounts of diverse data and prioritizing generality), for target identification, lead optimization, protein design, and clinical candidate selection. Drug discovery and development industry commentators have summarized the recent 10-15 years of history, distinguishing trends towards highly relevant data generation leveraged by generative AI and iterative feedback loops. Generative AI and foundation models at the molecular and tissue scales are openly available \citep{Takeda2023,Ingraham2023_chroma,Wohlwend2024_boltz,Chen2024,Hayes2024,Lu2024,M.Bran2024,batatia2023foundation}.
The question ``how to systematically leverage large language models and AI agents to coordinate and automate the action of agents in drug discovery and development?'' is not only raised, but is has received ambitious answers that that signal towards fuller automated decision making across the entire drug discovery and development pipeline \citep{swanson2024virtual}. However, integrating AI/ML into this hybrid scientific-strategic-business-regulatory process also presents challenges, such as data bias, overfitting, the multiple scales (length, time, energy) of measurements, the complex relationships between experimental and observational modalities, complex relationships between experimental modalities, lack of interpretability, and translation into actionable decisions. These challenges can be mitigated through strategies such as diverse data curation, simulation-based inference, explainable AI, and interdisciplinary validation \citep{Administration2023}.

While some voices emphasize a certain humility in the face of biological complexity \citep{Castaldo2024}, and a more discerning financing climate, others continue a positive narrative \citep{Chitnis2024}. One report, by authors from a Strategic Management Consulting firm ``that works with the world’s leading biopharmaceutical companies'', systematically documents successes \citep{KPJayatunga2024} through assessing the number of clinical assets in company pipelines. The authors looked into over 100 companies through an online databases that document the history of clinical assets, and followed up manually with finer grained desk research through press releases, company websites, \url{clinicaltrials.gov} and analyst reports. They find a substantially increased Phase I success rate over historical industry averages, and a comparable Phase II success rate.

Such topics formed the foundation of our roundtable discussion.

\paragraph{Discussion:}
We first summarize the professional biographies of the participants, and then summarize the discussion in sections on streamlining drug discovery, mitigating AI/ML integration risks, discovering novel biological knowledge, enhancing chemical design and synthesis, bridging science and strategy, before finishing with future directions.

Participants represented diverse expertise, from clinical areas like computational phenotyping, single cell RNA sequencing, digital pathology, functional molecular and clinical trials data integration, as well as the sub-nano scale of biomolecular simulation and structural biology. A group of about twelve were present during the full hour, and most had doctorates in the area of statistics, computer science, or some form of computational science and were mostly postdoctoral researchers or industry scientists; there were also a few doctoral students and one undergraduate.

\subsubsection{Streamlining Drug Discovery}  
The roundtable discussion touched on leveraging self-supervised learning, large-scale imaging, and high-dimensional phenotype modeling to decode biological systems and automate workflows. Participants opined that while measurable impacts include faster hit identification and integration of multimodal data, challenges remain in bridging biological complexity and technical scalability. Participants admitted current bottlenecks of data acquisition and a lack of experimental automation, but the discussion focused on the informal risk assessment and communication between ``computational" researchers and ``experimentalists" and associated communication challenges. Participants had deep expertise and technical focus, and perhaps because of this, the larger picture of how it all comes together in a successful pipeline remained largely elusive.   

\subsubsection{Mitigating AI Integration Risks}  
Participants noted risks like data bias, lack of interpretability, and slow validation cycles in drug discovery. Strategies to mitigate these include cross-disciplinary training to improve communication between machine learning practitioners and domain scientists, company culture surrounding decision making and risk aversion, using robust evaluation metrics, and adopting active learning approaches for iterative validation. The slow feedback loops in biological experiments amplify the difficulty, highlighting the need for well-grounded models and transparency in decision-making, and active learning perspectives like Bayesian optimization, where the cost of obtaining new data is itself modeled to optimize exploit/explore tradeoffs \citep{garnett_bayesoptbook_2022}.

\subsubsection{Discovering Novel Biological Knowledge} 
Participants commented on to what extent AI/ML has revealed new mechanisms, such as gene-gene or gene-drug interactions, from image and transcriptomics data. Digital twin models were highlighted as promising for simulating patient-level drug responses, albeit limited by challenges in verification and cross-scale integration. Simulating complex biophysical mechanisms at multiple scales, like how protein-ligand dynamics gives rise to cellular states, seems computationally intractable due to chaotic multi-scale interactions familiar to computational fluid dynamics. Despite this, this area is receiving attention, for example from the Chan Zuckerberg Institute for Advanced Biological Imaging's cryogenic electron tomography community data challenge \citep{Harrington2024} on a highly visible online platform. Participants discussed the pros and cons for hybrid approaches combining traditional physics-based methods with data-driven simulations and observations to advance understanding and do inference, and noted the NeurIPS 2024 workshop on Data-driven and Differentiable Simulations, Surrogates, and Solvers (\url{https://d3s3workshop.github.io/}).

\subsubsection{Enhancing Chemical Design and Synthesis}  
Participants offered anecdotes on to what extent AI/ML aids in computational compound generation and synthesis planning by predicting properties like solubility and toxicity, and to what extent this informs decision making criteria of domain expert experimentalists like medicinal chemists. Integration of simulation frameworks, akin to those in aerodynamics or chip design, was suggested to enable more real-time decision around drug prototypes. Furthermore, some natural language information like synthesis or even economic/fiscal aspects of regulatory reimbursement could perhaps benefit from large language model approaches.

\subsubsection{Bridging Science and Strategy}
The discussion emphasized fostering collaboration across disciplines and industries. Much of the discussion was taken up by anecdotal knowledge of how teams work together in different settings (startups, contract research organizations, clinical research centers, academic biomedical research institutes, internal consulting teams in large pharmaceutical companies). Participants appreciated learning from each other in a way that is not formally taught but typically gained through job experience and networking at meetings such as this one.

\subsubsection{Future Outlook} 
Participants suggested truly AI-native information-first teams either as an independent startup, or within larger pharmaceutical companies could accelerate progress with their agility and organization. Addressing cultural inertia and building credibility in AI/ML predictions were identified as crucial for broader adoption. Who will be the first to demonstrate never before seen success, which will drive cultural change within scientific teams?

\subsection{Social AI and Healthcare}


As artificial intelligence continues to evolve, its applications in healthcare have emerged as one of the most promising avenues for improving patient outcomes. Social AI, which focuses on understanding and interacting with human behavior, emotions, and social contexts, offers transformative potential for both mental and physical health interventions.  During this roundtable participants explored critical questions on the development of Social AI for healthcare:
How can ML-powered models of human social behavior improve the diagnosis and treatment of mental health conditions like depression, anxiety, and post-traumatic stress disorder (PTSD)? How do we combine and balance longitudinal history data and short-term behaviors for the modeling? How can Social AI enable real-time monitoring of patients' emotional and social health, both clinically and through wearable devices? How can Social AI tools adapt to individual patients' social and cultural contexts for equitable and effective care across diverse populations? How can models strike a balance between personalization and generalization for similar patients, benefiting from insights drawn from comparable cases?

\paragraph{Chairs:}
\textit{James M. Rehg, Yuwei Zhang, and Yurui Cao}

\paragraph{Background:} 

Advances in Machine Learning have led to the development of models capable of describing various aspects of human social communication and behavior.
These technological advancements hold significant potential for revolutionizing healthcare by introducing innovative methods for diagnosing and treating mental and developmental conditions, ultimately improving patient outcomes.

In the realm of mental health, AI's ability to model human behavior has been particularly impactful. Emerging tools now enable pre-diagnosis screening and risk assessment, offering insights into an individual’s predisposition to mental illnesses. For example, AI systems can analyze social media interactions to detect early signs of mental health disorders. Additionally, real-time monitoring through wearable devices and smart technologies enables continuous tracking of physiological and behavioral data, facilitating timely interventions while respecting patient autonomy.

Despite the promising applications, the integration of Social AI into healthcare presents challenges. Designing systems that adapt to individual patient needs and preferences is essential for effective care. Yet, personalization must be carefully balanced with concerns about bias and privacy to ensure equitable and secure healthcare delivery. Furthermore, incorporating social determinants of health into AI systems adds complexity, emphasizing the need for multidisciplinary approaches to ensure that these tools are both effective and inclusive.



    

\paragraph{Discussion:}
The discussion delved into AI's capacity to model human behavior, particularly in the areas of mental health diagnosis and treatment, and the management of developmental disorders such as autism and attention-deficit/hyperactivity disorder (ADHD). Participants explored how Social AI facilitates not only real-time emotional and behavioral health monitoring but also personalized interventions that adapt to individual patient needs. Tools such as wearable devices, smart home technologies, and mobile applications were highlighted for their ability to enable timely interventions while safeguarding patient privacy and autonomy. The roundtable also addressed the critical importance of personalization in AI-driven healthcare, emphasizing the balance needed between tailoring systems to individual patients and ensuring fairness, equity, and privacy across diverse populations. Discussions underscored the challenges posed by bias in healthcare data, the necessity of developing context-aware systems capable of addressing disparities, and the ethical responsibility to prioritize data privacy and transparent practices. These conversations highlighted both the immense potential and inherent challenges of integrating Social AI into healthcare, as participants envisioned a future where AI not only enhances our understanding of health but also drives more inclusive, accessible, and secure care.
\subsubsection{Human Behavior Modeling for Mental Health}

The discussion began with an example application focused on understanding textual communications between partners to track the frequency of conflicts.  Approaches brought up for modeling the mental health  in this context include calculating manually defined and statistically derived metadata features, extracting text embeddings using pretrained embedding models, as well as using modern large language models. Traditional ML approaches and LLMs could be used in conjunction to address the overall goal.

One persistent challenge in this area is inferring informative states that cannot be directly measured. While conditions like diabetes offer measurable indicators, mental health presents more difficulty in identifying clear, informative measures. This uncertainty underscores the challenge of determining what can be measured to accurately assess mental health, making it a key issue in the development of Social AI tools for healthcare applications.

\subsubsection{AI and Chatbots for Mental Health Support}

AI models, such as ChatGPT, are increasingly being used informally as therapeutic tools, where individuals engage in conversations to express frustrations and emotional experiences.  These interactions often elicit supportive and empathetic responses from the AI, which may seem overly idealized compared to human responses. While might be perceived as unusual initially, this practice highlights the non-judgmental nature of AI, which may appeal to users seeking a space free from human judgment. This raises questions about the potential role of AI in mental health support, especially in terms of its emotional neutrality and its capacity to provide a comforting, non-critical outlet for users.

While AI models like ChatGPT can provide supportive responses, they face limitations in recognizing and addressing urgent or complex situations. For example, if a patient experiences experiences financial misconduct, a human therapist might offer practical guidance, such as advising legal action. In contrast, AI may respond with general empathy and suggest continuing the conversation, failing to recognize the immediate need for legal intervention or other urgent support. 
This issue is compounded by safety concerns, as research has shown that AI models often struggle to follow instructions, such as avoiding trauma-triggering topics, which could lead to harmful outcomes, particularly in high-stakes mental health contexts.  This raises questions about their reliability and safety in emergency contexts.

\subsubsection{Real-Time and Longitudinal Monitoring through Wearable Devices}

Stress detection is an example application in the field of wearables for mental health. In such studies, participants wear devices that recorded physiological and accelerometry data. An algorithm could be used to predict stress levels, followed by a prompt asking participants to self-report their stress levels, providing a ground truth for validation. This approach reflects current practices in short-term mental health prediction, using wearable signals in conjunction with survey prompts to assess mental health. Incorporating long-term therapist notes into this data could provide valuable insights for further research. For example, the long-term clinical information could be a useful context for short-term predictions.

The discussion also highlighted the scarsity for long-termed data, especially in mental health (as it is not typically included in EHR data), along with the trade-off between eliciting accurate ground truth from the user and avoiding excessive prompting. While it is essential to prompt the user sufficiently to gather reliable data, over-frequent prompts may lead to participant fatigue or reduced engagement. On the other hand, insufficient prompting can result in a substantial amount of missing data, which may undermine the quality and completeness of the dataset. One approach to address this challenge is to utilize large language models for annotating and verifying the data, rather than directly labeling it.

Ethical concerns were also raised about companies mining user data without adequate consent, emphasizing the need for transparent data practices and privacy protection. Additionally, comparisons have been drawn between A/B testing and clinical trials, with A/B testing as a more flexible, real-world approach to experimentation, while clinical trials are held to stricter ethical and regulatory standards due to their focus on human health and well-being. It remains a key question how to balance these ethical considerations and practices when developing Social AI tools for healthcare applications.

\subsubsection{Personalized Healthcare through Social AI}

Personalization is a cornerstone of effective healthcare, and Social AI has the potential to tailor interventions to individual patients by considering their unique social, cultural, and behavioral contexts. Social AI tools can adapt to diverse populations by generating context-aware responses that align with patients' cultural norms and preferences. For instance, personalized health chatbots, such as those trained on individual user data, can provide consistent and long-term support, acting as virtual companions. These systems, designed to adapt to user behavior over time, exemplify the concept of ``personalized digital health companions," which remain relevant and effective for years.

One of the key challenges in personalization is balancing specificity with generalization. While personalized models can offer targeted recommendations by learning from an individual’s history, insights from similar patients can improve predictions and interventions for new users. This requires careful evaluation of the differences between patients and the identification of shared behavioral or contextual patterns. Context-aware models, designed to handle specific nuances of patient groups, are essential to addressing disparities in healthcare.

However, personalization introduces ethical challenges, such as ensuring predictive fairness across different demographic groups. Disparities in healthcare data can lead to biases in model performance, necessitating robust evaluation to guarantee equitable outcomes. Context-aware approaches, combined with rigorous testing across diverse populations, are critical to ensuring that personalized interventions benefit all patients without perpetuating inequities. Balancing privacy concerns with the need for individual data remains another challenge, highlighting the importance of transparent data practices and ethical guidelines in the development of Social AI tools.

\subsection{Population Health and Survival Analysis}

\paragraph{Subtopic:} 
How can we balance the complexity of ML-powered survival models with the need for interpretability in public health decision-making? Are some of the hybrid approaches helpful in retaining a level of transparency needed for public health decision-making?  What are the key challenges in validating ML survival models, such as random survival forests and deep learning approaches, on real-world administrative datasets? Would techniques such as nested cross-validation, simulation and external validation be helpful in understanding the performance of these approaches in real-world scenarios such as noisy, sparse and imbalanced data?  How can we design ML survival models that account for systemic biases in administrative data to ensure fairness in public health applications? Would incorporating socioeconomic and environmental data help adjust for health inequities and improve prediction accuracy for underrepresented groups? What practical steps are needed to overcome barriers to the adoption of ML methods in public health, such as computational intensity, lack of interpretability, and limited ML expertise among public health researchers? How can we encourage collaborations between data scientists and public health experts to co-develop models that align with public health goals?

\paragraph{Chairs: }
\textit{Mohammad Ehsanul Karim, Md. Belal Hossain, and Hanna A. Frank}

\paragraph{Background:}

Population health aims to understand and improve health outcomes across diverse groups by examining the complex relationship among biological, behavioral, environmental, and social determinants. In the context of survival outcomes, the focus extends beyond whether an event occurs and includes the timing of the event. Survival analysis, a statistical framework for modeling time-to-event data, is essential in public health research, enabling studies on disease progression, treatment effectiveness, and mortality. Among survival methods, the Cox proportional hazards model is widely used but has limitations. Its reliance on strong assumptions such as proportional hazards and correct model-specification often restricts its ability to capture the true relationships between predictors and outcomes.

In the modern era, nationally representative survey data and population-based administrative data have emerged as valuable resources for public health research. Nationally representative survey data, such as from National Health and Nutrition Examination Survey \citep{nhanes2024}, Canadian Community Health Survey \citep{cchs2024}, Korea National Health and Nutrition Examination Survey \citep{oh2021korea}, are publicly available and can be a valuable resource for identifying trends, associations, and building predictive models. On the other hand, health administrative data are derived from healthcare systems. These data cover entire populations or large cohorts, offering unique insights into health trends, disparities, and outcomes on a society-wide level. They provide longitudinal data spanning years or decades, facilitating comprehensive survival analyses and causal inferences over time. Unlike clinical trial data, survey data and health administrative data reflect real-world healthcare interactions and often include socioeconomic, demographic, and geographic variables critical for analyzing health disparities. The wide availability and cost-effectiveness of these data make it practical foundations for public health policies and resource allocation, such as identifying high-risk populations and addressing healthcare inequities.

Despite their potential, both survey and administrative datasets come with many challenges.  Survey data are generally serial cross-sectional in nature, some under- or over-sampling occurs, and samples may be dependent. Survey data generally include strata, clusters, and sampling weights to connect the sample to the population. Machine learning (ML) methods are mostly incapable of incorporating the strata, clusters, and weight variables. The health administrative datasets, on the other hand, include vast numbers of variables, such as diagnosis codes, prescriptions, and healthcare utilization, which complicate analyses; key variables are often incomplete, and imputation is difficult without robust parametric models; important predictors such as BMI or smoking status are often absent, leading to poor predictions or residual confounding for causal inference problems. Mismeasurement, inconsistencies, and coding errors (e.g., in ICD classifications) can undermine analysis; administrative data often includes time-dependent covariate and exposure information, which requires advanced methods to handle correctly; administrative data may reflect historical inequities in healthcare access and outcomes, which may not be addressed by statistical or ML methods.

ML methods have the potential to address some of these challenges and transform survival analysis. ML-powered survival models excel at handling high-dimensional data, capturing non-linear relationships, and mitigating residual confounding by incorporating additional information from the wealth of variables in the survey and administrative datasets. These methods also enable the development of actionable tools, such as risk calculators, to support public health decision-making and policy planning.

However, ML methods for survival analysis remain underutilized in public health. Key approaches, such as random survival forests, regularized survival models (e.g., Cox LASSO), and deep learning methods (e.g., DeepSurv, DeepHit) \citep{ishwaran2008random,katzman2018deepsurv,simon2011regularization,lee2018deephit}, show promise but face significant barriers to adoption. ML methods often lack the transparency of traditional models like Cox, which limits their acceptance among public health practitioners. Many ML methods are resource-intensive, particularly for large-scale datasets. These models are sensitive to hyperparameters, requiring additional diagnostics to ensure validity. Many ML methods are validated on curated datasets, and their performance on real-world data often noisy, sparse, and unbalanced – remains less established. ML models trained in survey or administrative data risk perpetuating systemic biases, and fairness metrics tailored to survival analysis are still underdeveloped. Additionally, not all ML approaches are equally suited for survival analysis. For instance, random survival forests are well-suited for capturing complex interactions while being moderately interpretable, but they can be computationally intensive for very large datasets. Deep learning methods, on the other hand, excel at modeling complex non-linear relationships and handling time-varying covariates or competing risks but often face challenges with interpretability and require substantial computational resources and large datasets for effective training.

To fully leverage the potential of ML-powered survival tools for public health, more research is needed to overcome these barriers. Advancing these tools will enable researchers to harness survey and administrative data effectively, improving health outcomes and equity on a population scale. By addressing challenges in implementation, validation, and fairness, ML-powered survival models can become indispensable in tackling the most pressing public health issues of our time.

\paragraph{Discussion:}
Participants discussed several solutions on the subtopics above.

\subsubsection{Balancing the complexity of ML-powered survival analysis with interpretability in public health.}
Data sources commonly used in public health research include survey data and health administrative data. Population surveys often use strategies like oversampling to create a sample including a large enough sample of underrepresented populations to enable subpopulation analyses. They also may use clustering strategies to reduce interview costs. To connect the sample back to the population, weights as well as strata and clustering information must be incorporated into the analysis. With ML methods like random forests and deep learning, weights are easily incorporated, but the inclusion of strata and cluster information is not as straightforward. 
Health administrative data, on the other hand, includes all those who interact with the healthcare system. In building clinical prediction models, it is generally recommended to restrict ourselves to the use of predictors that are helpful for the implementation and interpretation of the model. This often means relying on variables that are already established/known in the literature to be associated with the clinical outcomes of interest. However, health administrative data incorporates a vast amount of information, so restricting to those convenient for clinical settings ignores a large amount of useful information. For the creation of prediction models for use at a population level, such as for government use, it is worth considering what information from health administrative data we may be able to use to allow room for new discovery and the trade-off between the incorporation of these variables and the model’s interpretability.

Suggested possibilities for improving the interpretability of ML models in the public health context include the use of additive models. The development of software packages to facilitate this in survival analysis, including methods for feature selection, is currently in progress. The use of hybrid approaches, such as using investigator-specified predictors to create the base model and then trying to improve on this model’s predictive performance by including other variables from health administrative or claims data, was also suggested. Feature selection ML methods such as survival LASSO are also suggested since these methods can provide interpretable models with better accuracy. The possibilities of overfitting and underfitting of the ML-based models were also discussed. Participants emphasized the importance of balancing performance with interpretability, particularly in clinical contexts.

\subsubsection{Validation of ML survival models.}  
The thorough evaluation of survival models, using discrimination and calibration measures, was emphasized by participants as a vital step in ML-powered survival analysis. For discrimination, there was some debate as to which measure makes the most sense in the survival analysis context. Time-dependent area under the curve (AUC) is often used \citep{kamarudin2017time}, but this includes performance at very high false positive rates that may not truly be relevant in the application of a model. 

One of the difficult issues in survival modeling that came up in discussion was how to decide how and when subjects should be censored in the analysis. With longitudinal data, there are many situations in which people are lost to follow-up, whether because they are receiving care at different healthcare centers or moving between provinces/states or out of the country. How people should be censored is an important consideration in the design of the study, and with health administrative data it is usually impossible to guarantee we know what happens to every individual at every time point. Participants suggested the use of data sources like the Veterans Affairs (VA) data, in which generally everyone who receives care through VA receives all their care through VA. This way some of the issues with people receiving health care from different sources can be mitigated.

\subsubsection{Systemic bias in public health data sources and impact on ML methods.}  
One of the issues with using ML with health administrative data is that the ML methods can perpetuate systemic biases that are inherent in the dataset. For example, disparities in access to healthcare will be perpetuated by ML algorithm since those in marginalized populations will have less visits in the health administrative data. One proposed method to handle this is the creation of synthetic data, in which one creates a dataset that “oversamples” marginalized groups in order to create a dataset that more accurately represents the true population. However, such solutions may have their own issues (e.g., underestimation of variance, non-representativeness), so there is still a possibility that the data is biased.

\subsubsection{Practical steps for overcoming barriers to the adoption of ML methods in public health.} 
To address the barriers to the adoption of ML methods in public health, participants stressed the importance of group discussions and interdisciplinary collaborations. Limited domain knowledge prevents the wider use of ML in public health. Therefore, multidisciplinary discussions and the building of trust with ML and artificial intelligence programs are key steps to enabling the wider use of ML in public health. Additionally, the further development of comprehensive software infrastructure would make implementing these methods in various research areas easier.

\subsection{Personalization and Heterogeneity in Medicine}

\paragraph{Subtopic:} 
How can machine learning models effectively capture patient heterogeneity while ensuring personalized treatment recommendations?  What are the key challenges in balancing personalization with generalizability of solutions? How can we assess the generalizability/transportability of models that make personalized treatment recommendations? 

\paragraph{Chairs:}
\textit{Mohsen Sadatsafavi, Yuan Xia, and Sazan Mahbub}

\paragraph{Background:}

With the increasing availability of vast medical datasets, machine learning models offer great potential for delivering on the promise of Precision Medicine: enhancing clinical care by tailoring treatments to individual patient characteristics \citep{krishnan2023artificial}.

However, there are significant challenges along the way. Population heterogeneity threatens the transportability of models from one setting to another. Populations also change over time, leading to temporal shifts (aka calibration drift \citep{davis2017calibration}). Change in model performance when transported to a new setting may differ across subgroups, which can remain hidden and can aggravate existing disparities in care. Emerging treatments or changes in guidelines can disrupt predictions. One such disruptive change can be the adoption of the model itself, creating feedback loops between the model's adoption and the metrics used to evaluate its performance. Additionally, treatment assignments in the real world are not random and are influenced by factors such as the perception of disease severity, which are not always captured in the data \citep{kyriacou2016confounding}. Such 'confounding by disease severity', where patients with severe conditions are more likely to receive the treatment, can lead models to mistakenly associate effective treatments with poorer outcomes \citep{xia2024evaluating}.

Randomized controlled trials (RCTs) are the gold standard for evidence on treatment effectiveness in contemporary medicine and often a prerequisite for the approval of new treatments. However, RCTs are typically designed to estimate average treatment effect, while personalization focuses on conditional effect, whose estimation requires a significantly larger sample \citep{curth2024using}. Further, RCTs often have short follow-up periods, restrictive inclusion criteria, and strict study protocols that poorly reflect real-world practice. These limitations challenge many modern machine learning methods that rely on large, representative data.

The desire to create generalizable models might force developers to focus on features that are deemed to be resilient against distribution shifts, striking a balance between local accuracy and transportability \citep{subbaswamy2019preventingfailuresdatasetshift}. A similar trade-off might exist with respect to model explainability. Explainable models are more likely to be accepted by patients and care providers \citep{holzinger2019causability}, and this might incentivize model developers to focus on salient features. 

This round table was attended by individuals with interest in machine learning, causal inference, health policy, epidemiology, and biostatistics. This provided an opportunity to discuss, from diverse perspectives, how generalizability should be defined and measured, the challenges models face in predicting treatment effectiveness, and the trade-offs between making locally accurate predictions versus generalizable ones.

\paragraph{Discussion:}

To have a common ground for discussions, we decided to focus on machine learning  models that generate quantitative predictions about clinical outcomes. An early topic was the difference between predicting outcome risk versus treatment benefit. The former is akin to fitting a conditional mean function. The latter, on the other hand, is a `counter-factual' problem \citep{prosperi2020causal}. The group agreed that typical data that are used to train models are more suitable for predicting risk than predicting treatment benefit, due to the problem of unmeasured confounding (the summary from the Causality round table offers nuanced discussions). We noted that heterogeneity in treatment effect is scale-specific. The absence of treatment-by-feature interaction in one scale (e.g., relative risk) indicates its presence in other scales (e.g., absolute risk difference) \citep{greenland2009interactions}. It was mentioned that a common current practice is to use models to predict risk, to which the average treatment effect (e.g., relative risk reduction) from RCTs is applied to estimate the absolute risk reduction for a given patient \citep{Kent2020}. This approach inevitably ignores treatment-by-feature interactions on the relative scale. Participants noted that advances in causal machine learning based on both observational and RCT data have the potential to address these challenges \citep{feuerriegel2024causal}.

We debated whether certain models are more generalizable than others. Participants agreed that models are different in their generalizability, noting that generalizabiltiy can be formally considered at the development stage \citep{subbaswamy2019preventingfailuresdatasetshift}. However, it was mentioned that generalizability is a feature of both the domain (clinical condition of interest) and the model. For example, models that detect lung cancer on computed tomography images might be more transportable than models that detect individuals at high risk of self-harm based on their conversations with emergency staff. The latter is more likely to be affected by social norms, variations in practice standards, and so on. Participants agreed that clinical expertise in the domain of interest is key in the assessment of model generaliazability. One participant suggested explicit modeling of population heterogeneity for a model via network meta-analysis methods, which enables making probabilistic statements about the performance of the model in a new population \citep{Phillippo2020NMA}.

On the relationship between personalization and generalizability, we discussed that they do not necessarily move in opposite directions (indeed, a perfect oracle will make ultimately accurate and generalizable predictions). However, we acknowledged that different features in the data might vary in their susceptibility to distribution or domain shifts, independently of their predictive power. Participants were asked about the notion that high-dimensional models might be more generalizable, as such models might capture variables that are responsible for differences across populations. A good example of this concept is foundation models \citep{bommasani2021opportunities}. These models are trained on vast amounts of data spanning one or multiple domains, enabling them to learn generalizable priors. As a result, foundation models are expected to exhibit broad applicability, demonstrating the ability to generalize across diverse tasks and settings. However, this was challenged by the counter-argument that such models also have a tendency to fit to local features (or miss local features in a new setting that determine the outcome), leading to significant performance degradation when transported~\citep{kernbach2022foundations, tirumala2022memorization}.

We made a distinction between generalizbility of models in terms of discrimination and calibration. Research in clinical prediction modeling has shown that when models are transported to new populations, their calibration (e.g., change in calibration slope) is often substantially affected, more so than their discrimination (e.g., change in AUC) \citep{van2019calibration, gulati2022generalizability}. This led to the discussion that models can be more generalizable when they are used for ranking (e.g., which three hospitalized COVID-19 patients are at the highest risk of deterioration and should use the three ventilators in the ward?). In contrast, when numerical predictions are communicated (e.g., your risk of heart attacks in the next 10 years), much emphasis should be placed on calibration. 

Towards the end, we discussed different ways models can be evaluated (validated) in a new population. While we universally agreed that models need to be tested before deployment, we also noted that for models that hold promise, waiting for the results of the validation study can itself cause opportunity loss (especially for predictions over long time-horizons, where a validation study might take years). As such, in some circumstances, one might conditionally deploy a model while learning about its performance. Stepped wedge designs were mentioned to be suitable for simultaneous implementation and learning \citep{kappen2018evaluating}. A related point was made on the extent one should fine-tune a model based on available information before a full-on validation study, such as accounting for differences in average outcome risk between training and target populations \citep{sadatsafavi2022marginal}. However, there were concerns about the misuse of such `blanket' corrections. 

Overall, participants highlighted the interdisciplinary expertise required for developing and deploying models that are precise and generalizabile, and at the same time acceptable to patients, providers, and regulators. There was a broad consensus that this is a context-specific task, and any one-size-fits-all recommendation might not itself be generalizable.

\section{Summary}

The ML4H 2024 Research Roundtables highlighted diverse challenges and opportunities in the application of machine learning to healthcare. Among the most discussed topics was the application of multimodal foundation models capable of processing diverse data types. However, significant challenges remain in handling incomplete data, varying data quality, temporal alignment, and maintaining privacy across modalities. The discussions highlighted the importance of understanding clinical context and sampling biases when developing these models.

Causality emerged as a critical consideration. Participants noted the need for standardized benchmarks, evaluation protocols, and reporting guidelines. Causal models can offer increased robustness, however there can be a trade-off between causal validity and predictive performance. The discussions emphasized the value of developing ways to disentangle causal and correlational representations in healthcare models.

Data standardization and benchmarking pose persistent challenges. Despite the availability of public healthcare datasets, researchers struggle with preprocessing standardization, reproducibility, and clinical validation of variables. For low- and middle-income countries, unique challenges exist in implementing healthcare AI, including limited infrastructure and data availability. Participants discussed potential solutions including synthetic data generation, robust partnerships between developed and developing regions, and the importance of stakeholder engagement. The discussions emphasized the need to prioritize data quality and representativeness while ensuring solutions are contextually appropriate and ethically sound.

Participants across sessions noted how AI systems can perpetuate existing biases and exacerbate disparities in care. The roundtable on fairness and bias highlighted the need for epistemic humility and meaningful inclusion of diverse perspectives in AI development, including input from social scientists, bioethicists, and patients. Participants expressed concern about structural barriers to inclusive research, and extractive and exploitive AI practices. The discussion highlighted novel methodological approaches to measuring disparities and called for educational and structural reform in ML and medicine. 

A recurring theme was the need for thoughtful integration of AI into existing clinical workflows, with participants highlighting the gap between research and deployment. Most evaluated models are not deployed, but many deployed models are not evaluated. Participants were particularly concerned about regulation loopholes that allow EHR providers to deployed models without demonstrated clinical utility, proper monitoring protocols, and governance structures. Consistent with emerging regulatory frameworks in Europe and around the world, participants in several roundtables emphasized that different AI tools require different levels of oversight depending on their risk levels and potential for harm.

Rather than pursuing AI as a solution in itself, participants emphasized the importance of understanding real clinical needs and contexts. This includes carefully considering whether one should even consider AI solutions for a particular problem, and how we can adapt tools to specific context and varying healthcare resource settings. The discussions also highlighted the importance of continuous evaluation and monitoring of AI systems after deployment, particularly given the complex and evolving nature of healthcare environments.

The interdisciplinary nature of ML4H research requires on strategic partnerships and continuous proactive effort and learning to bridge the knowledge gap. Success often depends on \textit{bilingual} experts in the team who can effectively translate between technical and clinical domains to facilitate better communication and innovation. Rotations that provide clinicians and computer scientists with first-hand experience of the realities and challenges faced by the other field can be very beneficial. 

Finally, the conversations underscored the importance of engaging stakeholders, earning trust, and ensuring equity in AI-driven healthcare solutions, particularly in low-resource settings.



\section{Lessons Learned}

Overall, recruiting senior and junior chairs went smoother compared to the previous year, except for niche interdisciplinary areas such as health economics. Limited funding for attending the conference remains an issue for many potential chairs. Delays in visa processing, which unfairly affect participants from lower- and middle-income countries, prevented some participants from attending the conference. 

This year, the roundtable chairs came from diverse academic and industry backgrounds, which enriched the discussions. The longer 50-minute format allowed for more in-depth discussion and networking among participants. Overall, roundtable sessions were well-attended and popular among participants and chairs. However, the large number of participants who attended research roundtables which were held on another floor in the venue caused significant congestion beyond elevator capacities. 

After the conference, junior and senior chairs provided summaries of discussions, which was used to create this document. We recommend ML4H 2025 conference to continue incorporating research roundtables. Introducing variations and additional focus to topics each year, which could be inspired by local interests and expertise, can help ensure research roundtables remain engaging and relevant for years to come. 

\clearpage

\bibliography{references}

\end{document}